\documentclass{article}

\usepackage{iclr2027_conference,times}

\usepackage{amsmath,amsfonts,bm}

\def\eqref#1{equation~\ref{#1}}

\def\1{\bm{1}}

\DeclareMathAlphabet{\mathsfit}{\encodingdefault}{\sfdefault}{m}{sl}
\SetMathAlphabet{\mathsfit}{bold}{\encodingdefault}{\sfdefault}{bx}{n}

\usepackage{amsmath}
\usepackage{booktabs}
\usepackage{listings} 
\usepackage{wrapfig} 
\usepackage{flafter}
\usepackage{graphicx}
\ifdefined\ifparticipantjourneys\else
  \expandafter\newif\csname ifparticipantjourneys\endcsname
  \participantjourneystrue
\fi
\usepackage{hyperref}
\usepackage{xurl} 
\usepackage[normalem]{ulem}
\usepackage{iftex}
\newcommand{\reviewcomment}[1]{}
\providecommand{\reviewposition}[2]{}
\usepackage{tikz}
\usepackage{pgfplots}
\pgfplotsset{compat=1.18}
\usepgfplotslibrary{groupplots}
\usepackage{placeins}

\definecolor{kstext}{HTML}{2D2D2A}
\definecolor{kstextmuted}{HTML}{7A7A72}
\definecolor{ksaccentA}{HTML}{E07A5F} 
\definecolor{ksaccentB}{HTML}{81B29A} 
\definecolor{ksaccentC}{HTML}{F2CC8F} 
\definecolor{ksaccentD}{HTML}{B0A8C8} 
\definecolor{ksborder}{HTML}{E0DED8}

\pgfplotsset{
  earthtone/.style={
    axis line style={color=ksborder, line width=0.6pt},
    tick style={color=kstextmuted, line width=0.5pt},
    tick label style={color=kstextmuted, font=\scriptsize},
    label style={color=kstext, font=\scriptsize\bfseries},
    title style={color=kstext, font=\small\bfseries, yshift=-2pt},
    legend style={fill=none, draw=none, text=kstext, font=\scriptsize,
                  inner sep=0pt, row sep=1pt,
                  /tikz/every odd column/.append style={column sep=1pt}},
    legend cell align=left,
    grid=none,
  },
  ksfill/.style={
    only marks, mark=*, mark size=2.2pt,
    mark options={fill=#1, draw=#1!55!black, line width=0.5pt},
    error bars/.cd, x dir=both, x explicit,
    error bar style={color=#1!55!black, line width=0.6pt},
    error mark options={rotate=90, mark size=1.5pt, color=#1!55!black},
  },
  kshollow/.style={
    only marks, mark=*, mark size=2.2pt,
    mark options={fill=white, draw=#1!70!black, line width=0.9pt},
    error bars/.cd, x dir=both, x explicit,
    error bar style={color=#1!70!black, line width=0.6pt},
    error mark options={rotate=90, mark size=1.5pt, color=#1!70!black},
  },
  ksvline/.style={
    color=#1!60!black, line width=0.7pt, mark=*, mark size=2pt,
    mark options={fill=#1, draw=#1!55!black, line width=0.5pt, solid},
    error bars/.cd, y dir=both, y explicit,
    error bar style={color=#1!55!black, line width=0.5pt, solid},
    error mark options={mark size=1.4pt, color=#1!55!black},
  },
  ksvfill/.style={
    only marks, mark=*, mark size=2.2pt,
    mark options={fill=#1, draw=#1!55!black, line width=0.5pt},
    error bars/.cd, y dir=both, y explicit,
    error bar style={color=#1!55!black, line width=0.6pt},
    error mark options={mark size=1.5pt, color=#1!55!black},
  },
}

\colorlet{kshuman}{ksaccentA}   
\colorlet{ksdirect}{ksaccentB}  
\colorlet{ksharness}{ksaccentD} 
\usepackage{tcolorbox}
\tcbuselibrary{breakable}
\newtcolorbox{participantinstructions}[1]{
  breakable,colback=black!2,colframe=black!45,boxrule=0.5pt,
  arc=1mm,left=8pt,right=8pt,top=6pt,bottom=6pt,
  title={#1},fonttitle=\bfseries,colbacktitle=black!7,coltitle=black}

\usepackage{xcolor}
\newif\ifpaperrevisions
\paperrevisionsfalse

\newcommand{\arc}{ARC-AGI-3}
\newcommand{\vista}{\textsc{Vista}}

\newcommand{\dat}[1]{#1}

\title{How Humans Learn Unfamiliar Worlds:
A Think-Aloud Dataset on ARC-AGI-3}

\author{Shuzhi Liu \\
Nanyang Technological University\\
\texttt{SHUZHI001@e.ntu.edu.sg}
\And
Archana Warrier \\
TU Darmstadt\\
\texttt{archana.warrier@tu-darmstadt.de}
\AND
Dat Nguyen \\
Basis Research Institute \& Harvard University\\
\texttt{dat@basis.ai, datnguyen@seas.harvard.edu}
\And
Chengqi Zheng \\
Nanyang Technological University\\
\texttt{CZHENG014@e.ntu.edu.sg}
\AND
Kevin Ellis \\
Cornell University\\
\texttt{kellis@cornell.edu}
\And
Yewen Pu \\
Nanyang Technological University\\
\texttt{yewen.pu@ntu.edu.sg}
}

\iclrfinalcopy  

\begin{document}

\maketitle
\lhead{Preprint} 

\begin{abstract}
Learning through exploration and using newly acquired knowledge to achieve
goals are central capabilities of general intelligence. Understanding these
capabilities requires tracing how learners build, use, and revise world models
during interaction.
We construct the first dataset that records how humans discover and use the mechanisms of
novel
environments, using ARC-AGI-3 as a testbed: 144 trajectories from 64 participants across 25 public games, each pairing time-synchronized actions with think-aloud audio, plus 250 trajectories from large language model (LLM) agents with and without harnesses on the same games.
We annotate mechanism triggers and recognition, action purposes, and errors
and recovery, linking changes in understanding to the observations and actions
that support them.
We analyze characteristics of human cognition and compare them with those
of LLM agents. Both groups anticipate mechanisms and use
familiar analogies. Qualitative representations are more common in successful
human trajectories than in successful LLM trajectories (50.0\% versus 14.3\%).
Humans more often explore unfamiliar situations without a concrete hypothesis,
while hypothesis testing is common in both groups. We also trace how learners
recover through belief revision, replanning, and execution correction.
As LLMs with well-designed harnesses reach saturated performance on public
ARC-AGI-3 games, our cognitive analyses motivate next-generation benchmarks
that challenge familiar priors, require accumulated evidence and abstraction,
and test planning through subgoals, experimentation, and recovery.
Our dataset is available at \url{https://arc3-human-think-aloud.github.io/}.
\end{abstract}

\section{Introduction}
\label{sec:introduction}

Discovering and using unfamiliar rules is central to general
intelligence~\citep{legg2007universal,chollet2019measure}. Actions both reveal
and change an environment, requiring learners to gather information while
pursuing goals. Unexpected outcomes can prompt experiments, revised rules,
or new plans. Humans learn new video games within
minutes~\citep{tsividis2017human,Lake2016building}, solve physical puzzles in
few attempts~\citep{allen2020rapid}, and experiment with unfamiliar toys even
as children~\citep{cook2011where,Schulz2012origin}. Interactive benchmarks
compare agents with humans on success and action
efficiency~\citep{arcprize2026arcagi3,warrier2025worldtest}, with large language
model (LLM) agents rapidly closing the gap~\citep{han2026vista,zeng2026schema}.
These scores leave cognitive processes largely unobserved. Studying them
requires tracing what learners understand as they explore and act.

Existing human data rarely reveal what learners understand at each step.
Action traces alone are ambiguous: repeatedly pushing an object could reflect
experimentation, executing a solution, or persisting under an incorrect belief.
ARC-AGI-1 studies~\citep{chollet2019measure} record solutions, editing traces,
rule descriptions, and instructions~\citep{johnson2021fast,legris2025harc,acquaviva2022larc,kim2025arctraj},
but concern generalizing from examples rather than exploring an environment.
The \arc{} human release records actions but not their underlying
reasoning~\citep{kamradt2026humandataset}. Think-aloud datasets pair reasoning
with behavior~\citep{wurgaft2025thinkaloud,liu2026egotl}, but do not require
discovering both unfamiliar dynamics and goals. Studying this process requires
concurrent reports aligned with actions and environmental responses.

We construct the first dataset tracing how humans discover and use the
\emph{mechanisms} of novel interactive environments
(Figure~\ifparticipantjourneys\ref{fig:participant-journeys}\else\ref{fig:method-overview}\fi).
Our testbed, \arc{}, requires players to infer mechanisms and goals through
interaction without instructions~\citep{arcprize2026arcagi3}. Its games rely on
core object and spatial priors; successive levels introduce or recombine
mechanisms. Using concurrent think-aloud~\citep{ericsson1993protocol}, we
collect 144 successful trajectories from 64 participants across 25 public games,
with 136,876 actions aligned with speech, including errors and recovery.
We also collect 250 public LLM trajectories on these games: 100 from direct
models and 150 using \vista{}, Schema, OPINE-World, and Astra
harnesses~\citep{han2026vista,schema2026traces,opine2026,arcprize2026astraresults}.

Our annotation framework covers three components:
world-model changes, action generation, and errors and recovery
(Section~\ref{sec:method}). We catalog 354 game mechanisms, locate their triggers
and recognition, classify the purposes of action segments, and link errors to
subsequent evidence of correction. This alignment supports examining how
understanding develops and guides action throughout a task.

Section~\ref{sec:human-world-models} examines prior knowledge, qualitative
representations, action purposes, and errors and recovery. Both humans and
LLMs anticipate mechanisms and use familiar analogies. Qualitative descriptions
occur in 50.0\% of 144 successful human trajectories versus 14.3\% of 119
successful \vista{}, Schema, and OPINE-World trajectories. Both groups act
to pursue goals, gather information, or both. Humans more often explore
without a concrete hypothesis; hypothesis testing is common in both groups.
Recovery involves belief revision, replanning, and execution correction:
11 of 15 incorrect mechanism conclusions in human reports and 451 of 593 in
successful LLM records are subsequently corrected. As well-designed harnesses
saturate performance on public \arc{} games, these analyses motivate benchmarks
that challenge familiar priors, require accumulated evidence and abstraction,
and test planning through subgoals, experimentation, and recovery
(Section~\ref{sec:implications}).

Our main contributions are as follows:
\begin{itemize}
  \item \textbf{Dataset.} We release the first dataset of concurrent human
  think-aloud reports aligned with actions and environmental feedback in
  environments whose mechanisms and goals learners must discover, together with
  matched LLM agent trajectories and mechanism-level annotations of
  triggers, recognition, action purposes, errors, and corrections.
  \item \textbf{Analysis of humans and LLM agents.}
We use aligned records to examine prior knowledge,
  qualitative world-model representations, action purposes, and errors and
  recovery, comparing human and LLM evidence and identifying questions for
  further study.
  \item \textbf{Implications for benchmark design.} We propose
  benchmarks with mechanisms that defy familiar priors, require cumulative
  induction, and extend planning through subgoals and recovery.
\end{itemize}

\section{Related Work}
\label{sec:related-work}

We review two lines of work: (1)~datasets of human behavior and reasoning,
which lack concurrent reports from people exploring unfamiliar environments;
and (2)~\arc{} and its agent harnesses, which provide a setting for collecting
such reports and public LLM trajectories to compare them against.

\paragraph{\dat{Human behavioral and reasoning-process datasets}} Think-aloud protocols provide evidence about reasoning during problem
solving~\citep{ericsson1993protocol}. Recent work scales their transcription
and annotation in the Game of 24~\citep{wurgaft2025thinkaloud}, uses verbal
reports to constrain cognitive models of risky choice~\citep{xie2026thinkaloud},
and compares human and LLM-generated reasoning in chemistry tutoring
\citep{borchers2026students}. EgoTL aligns spoken reasoning with egocentric
video and actions in household tasks~\citep{liu2026egotl}.

%
ARC-AGI-1 tasks require learners to infer grid transformations from examples
and apply them. Studies including H-ARC collect human solutions, action traces,
and rule descriptions~\citep{johnson2021fast,legris2025harc}; LARC captures
language used to communicate solutions~\citep{acquaviva2022larc}, and ARCTraj
records object-level editing trajectories~\citep{kim2025arctraj}.

%

These resources cover mathematical problem solving, choices between specified
options, chemistry exercises, everyday household activities, and abstract
grid transformations. They do not provide concurrent reports of humans
discovering both the dynamics and goals of unfamiliar interactive environments.
Our dataset fills this gap by recording thoughts as people explore such
environments, aligning their evolving hypotheses and plans with the actions
and feedback that inform them.

\paragraph{\dat{ARC-AGI-3 and agent harnesses}}
\arc{} provides a suitable setting for studying how people explore and learn
an unfamiliar environment: players must discover its mechanisms and goals through
interaction~\citep{arcprize2026arcagi3,arcprize2026arcagi3overview}. Building on ARC's emphasis on skill
acquisition~\citep{chollet2019measure}, it presents colored grids that players
interact with through key presses or clicks. Players infer the mechanisms
and completion conditions from visual feedback without natural-language task
instructions. The games build on basic priors such as objects and space
and require no specialized knowledge. Each game comprises multiple levels,
which may introduce new mechanisms or require combining known mechanisms
in a new way. Progress therefore involves discovering relevant mechanisms
and applying them to reach each level's completion condition; the benchmark
measures performance relative to human action efficiency~\citep{arcprize2026arcagi3}.
We refer to individual games by their identifiers, such as BP35 and TU93.

The official \arc{} human release provides gameplay replays in environments
with unknown dynamics~\citep{kamradt2026humandataset}.
These replays record actions and outcomes but do not include concurrent
verbal reports of the hypotheses, expectations, and plans behind them.
We collect these reports to study how people discover and use the
environment's mechanisms.

Recent work designs harnesses that achieve strong performance
on the public \arc{} environments~\citep{han2026vista,zeng2026schema}.
Their designs support memory, model construction, and planning: \vista{}
stores retrievable visual experience~\citep{han2026vista}, Schema constructs
executable models~\citep{zeng2026schema}, and OPINE-World combines hypothesis
testing, model synthesis, and replay verification~\citep{courtis2026opine}.
We collect the LLM trajectories and reasoning records that have been 
released publicly, including those from \vista{}, Schema, and OPINE-World, to compare how LLMs and humans
form world models, choose actions, and correct errors during interaction
\citep{han2026vista,schema2026traces,opine2026}.

\section{Method}
\label{sec:method}
\dat{In this section, we describe our method for dataset construction towards the goal of exploring the process of dynamically learning world models in humans and agents. We introduce the preliminaries on world model learning in Section~\ref{sec:preliminaries}, describe data collection in Section~\ref{sec:data-collection}, and present the data annotation process in Section~\ref{sec:data-annotation}.}

\subsection{Preliminaries}
\label{sec:preliminaries}

%

Interactive learning in unfamiliar environments is a central component of general intelligence~\citep{chollet2019measure,ying2025assessing}. Throughout interaction, learners build and revise a \emph{world model}: their understanding of the environment's objects, relations, dynamics, and goals. This understanding guides the actions they take, whose outcomes provide further evidence for
 updating it~\citep{warrier2025worldtest}. We describe this recurring process as
\begin{equation}
  (a_t,W_{t+1})=\operatorname{Think}(W_t,\mathcal{T}_{\leq t}),
  \label{eq:world-model-think}
\end{equation}
where $W_t$ is the current world model, $\mathcal{T}_{\leq t}$ contains the
observations and action--outcome history available before action $a_t$, and
$W_{t+1}$ is the revised model. The action produces a new observation that
enters the next round of thinking.

To analyze these cognitive processes, following~\citet{zhou2025walle2},
we represent a world model as a list of mechanisms $W=(m_1,\ldots,m_K)$.
Each mechanism is a conditional state-update rule:
\[
  m:\quad C_m(x_t,a_t)=1\ \Longrightarrow
  x_{t+1}[V_m]=E_m(x_t,a_t).
\]

\definecolor{mechbg}{HTML}{F3F8FD}   
\definecolor{mechrule}{HTML}{D6E4F3}
\definecolor{mechkw}{HTML}{2E7D32}    
\lstdefinelanguage{mechanism}{
  morekeywords={if,and,then},
  sensitive=true,
}
\lstdefinestyle{mechanism}{
  language=mechanism,
  basicstyle=\fontencoding{T1}\fontfamily{zi4}\fontsize{6}{7.4}\selectfont\color{kstext},
  keywordstyle=\bfseries\color{mechkw},
  emph={M1,M2,M3,M4},
  emphstyle=\bfseries\color{kstext},
  columns=fixed,
  basewidth=0.5em,
  upquote=true,
  literate={-}{{-}}1,
  keepspaces=true,
  breaklines=true,
  breakindent=4.2em,
  backgroundcolor=\color{mechbg},
  frame=tb,
  rulecolor=\color{mechrule},
  framesep=4pt,
  xleftmargin=4pt,
  framexleftmargin=4pt,
  xrightmargin=7pt,
  aboveskip=0pt,
  belowskip=0pt,
}

\begin{wraptable}{r}{0.35\textwidth}
\vspace{-22pt}
\caption{Mechanisms from BP35.}
\label{tab:mechanism-examples}
\vspace{4pt}
\begin{lstlisting}[style=mechanism]
M1 if   left pressed and left cell clear
   then player moves one cell left
M2 if   left pressed and solid block left
   then player's x-position is unchanged
M3 if   gravity up and cell above clear
   then player moves up
M4 if   green support clicked
   then support is removed
\end{lstlisting}
\vspace{-14pt}
\end{wraptable}

Here, $x_t$ is the full game state and $a_t$ is an action. The condition
$C_m$ returns 1 when the mechanism applies and 0 otherwise. The effect function
$E_m$ outputs the new values of the state variables $V_m$ affected by the
mechanism; $x_{t+1}[V_m]$ denotes those variables in the next state.
%
A mechanism specifies a local update; multiple mechanisms jointly determine
the transition. Table~\ref{tab:mechanism-examples} gives natural-language
examples of mechanisms from the game BP35.

%

\par
\ifnum\value{WF@wrappedlines}>1
  \vspace{\dimexpr\value{WF@wrappedlines}\baselineskip-\baselineskip\relax}
\fi
\WFclear

\subsection{Data collection}
\label{sec:data-collection}

We use the concurrent think-aloud method~\citep{ericsson1993protocol}
to collect people's actions and their thoughts during gameplay.
We recruit paid participants through Prolific.
Before the formal tasks, participants read the study instructions, provide
informed consent, complete a microphone check, and receive a tutorial on the
interface controls.
%
They then complete an attention-check level while thinking aloud;
passing this check is required to enter the formal study.
%
%
Each participant is asked to complete three games.
Participants explore the unknown mechanics and goals while continuously
verbalizing their thoughts as they attempt to win. Instructions emphasize
detailed verbalization over rapid completion, and an on-screen reminder prompts
participants to continue speaking after %
a period of silence.
We record game actions and audio concurrently, preserving behavior and verbal
reports during exploration, mechanism learning, and plan execution.
Appendix~\ref{app:human-protocol} provides the instructions, time limits,
and further protocol details.

%
The human corpus comprises 144 successful trajectories from 64
participants across 25 public \arc{} games, containing 136,876 environment
actions with concurrent verbal reports.
%
\ifparticipantjourneys
Figure~\ref{fig:participant-journeys} shows three collected human trajectories
from the first levels of LS20, BP35, and FT09, including the participants' attempts,
revisions, and eventual solutions. Each trajectory pairs gameplay frames with
concurrent think-aloud excerpts by action step.
\else
Figure~\ref{fig:method-overview}(a) shows an example of our collected human
trajectories and concurrent think-aloud reports, aligning gameplay frames and
verbal reports by action step.
\fi
These paired records support cognitive analyses
of how people understand unfamiliar environments, choose actions, and solve
interactive tasks.\reviewcomment{E01}

\ifparticipantjourneys
\begin{figure}[!t]
  \centering
  \includegraphics[width=\textwidth,height=.79\textheight,keepaspectratio]{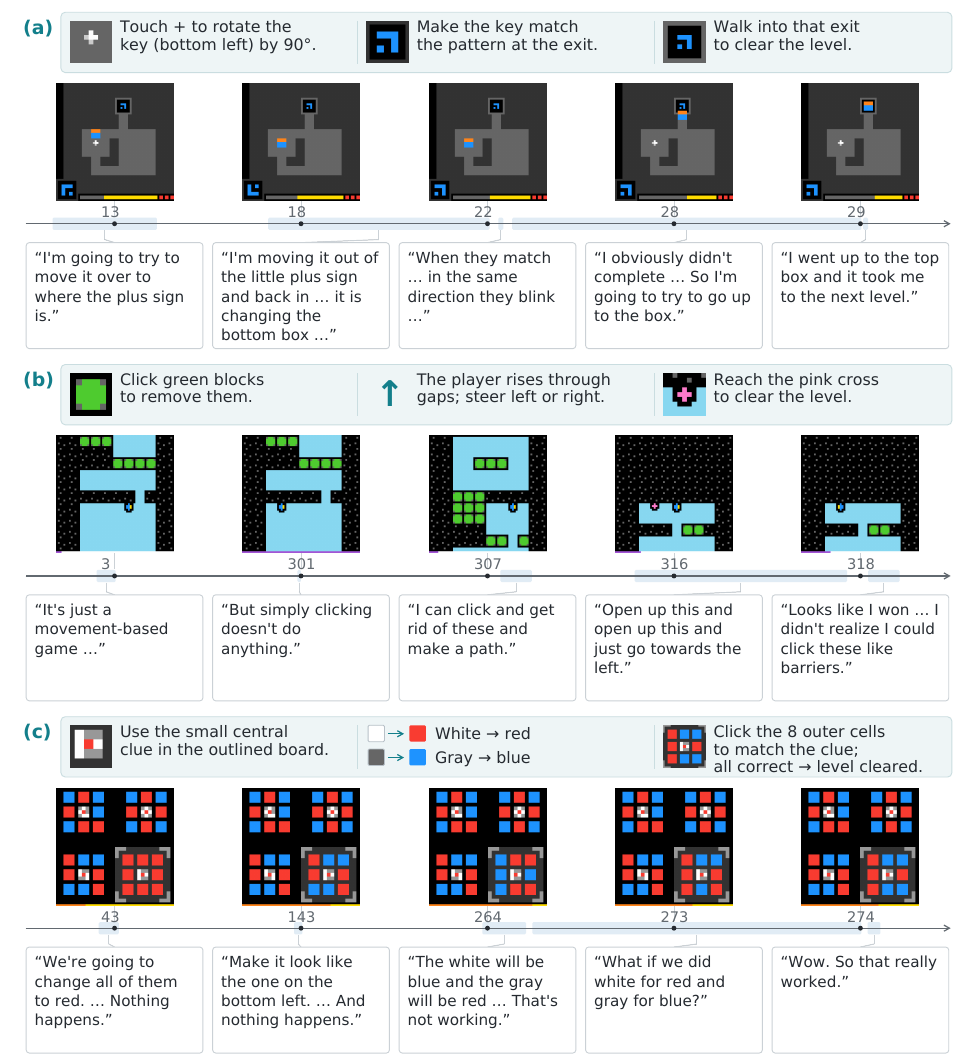}
  \caption{\textbf{Human trajectories and concurrent think-aloud data.}
  Key frames and aligned think-aloud excerpts trace participants' attempts to solve
  the first level of (a) LS20, (b) BP35, and (c) FT09.
  Each row includes the key mechanics needed to solve that level.}
  \label{fig:participant-journeys}
\end{figure}
\fi

We also collect publicly released action trajectories and available reasoning records from different LLMs, both in
direct configurations and with harnesses including \vista{}, Schema,
OPINE-World, and the Astra Provider Adapter, to compare human and LLM
exploration, world-model induction, and planning processes
\citep{han2026vista,schema2026traces,opine2026,arcprize2026astraresults}.
This corpus contains 250 trajectories across the same 25 games:
100 from direct agents and 150 from agents with harnesses, with one
trajectory per game for each configuration.
System configurations and comparison scope are detailed in
Appendix~\ref{app:experiment-details}.\reviewcomment{A09}


\subsection{Data annotation}
\label{sec:data-annotation}

%

We annotate three aspects of human and LLM interaction: the cognitive
processes involved in forming a world model, the purposes of actions, and
errors and recovery. Human think-aloud reports and public agent records are
aligned with game actions and observations, so that each annotation can be
examined in its interaction context (Figure~\ref{fig:method-overview}).

\ifparticipantjourneys
\begin{figure}[!htbp]
  \centering
  \includegraphics[width=\textwidth]{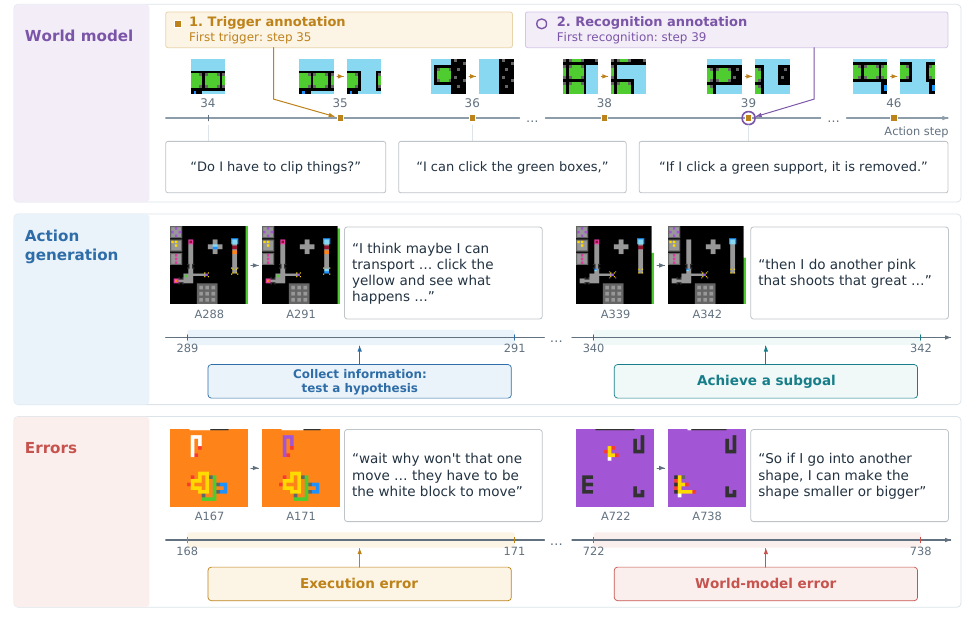}
  \caption{\textbf{Data annotation.} Annotations cover three components:
  \emph{world-model changes}, tracking mechanism triggers and when and how learners
  recognize mechanisms; \emph{action generation}, distinguishing goal pursuit,
  information gathering, and actions serving both purposes; and \emph{errors and
  recovery}, identifying mechanism, planning, and execution errors and linking
  them to any subsequent corrections.}
  \label{fig:method-overview}
\end{figure}
\else
\begin{figure}[!htbp]
  \centering
  \includegraphics[width=\textwidth]{figures/method_overview/method_overview_v13.pdf}
  \caption{\textbf{Data collection and annotation.}
  (a) An example of the collected data, comprising a human gameplay trajectory
  and its aligned, concurrent think-aloud transcript.
  (b) Annotations cover three components: \emph{world-model changes}, tracking
  mechanism triggers and when and how learners recognize mechanisms;
  \emph{action generation}, distinguishing goal pursuit, information gathering,
  and actions serving both purposes; and \emph{errors and recovery}, identifying
  mechanism, planning, and execution errors and linking them to any subsequent
  corrections.}
  \label{fig:method-overview}
\end{figure}
\fi

\subsubsection{Changes in the world model}
\label{sec:world-model-annotation}
\label{sec:mechanism-annotation}
\label{sec:trajectory-annotation}

\paragraph{Game mechanism annotation.}
We manually annotate the mechanisms designed into each \arc{} game.
Two authors jointly annotate the mechanisms through discussion, describing
each mechanism in natural language.\reviewcomment{A10}\reviewcomment{D08}
Some mechanisms are not required for game completion, such as failure
conditions and boundary conditions on the agent's maximum movement range.
Across 25 games, the catalog contains 320 mechanisms required for completion
and 34 optional game mechanisms.

\paragraph{Mechanism trigger annotation.}
For each trajectory, we annotate when each mechanism is triggered: the times
at which its conditions are satisfied and it affects the next state.
We insert trigger-detection code into the \arc{} game source. The detectors
check the values and changes of mechanism-specific variables before and after
actions, as well as whether relevant code paths execute, to locate mechanism
triggers. We use GPT-5.6 Sol to implement the detectors, and one author
spot-checks the resulting annotations. All 30 manually checked trigger
annotations were correct.\reviewcomment{D09}

\paragraph{Cognitive changes in the world model.}
We infer changes in people's world models from their trajectories and
think-aloud reports. Participants sometimes describe changes in their
understanding in detail; at other times, they express an insight only through
an interjection, in which case we infer their understanding indirectly from
subsequent actions. We provide GPT-5.6 Sol with timestamped transcripts and
trajectory data and ask it to infer changes in the world model wherever the
available evidence permits. Because participants do not verbalize every
change in their thinking, their understanding of some mechanisms remains
unclear. In particular, we annotate the first time each mechanism is correctly
recognized and the participant's description of that mechanism. Recognition
can precede or follow the first trigger.\reviewcomment{A11}
We manually reviewed 100 recognition annotations; 97 were correct
(97\% accuracy).\reviewcomment{A12}

\subsubsection{Action generation}
\label{sec:action-purpose-annotation}

We annotate the purposes of contiguous action segments using the associated
verbal or reasoning records and the surrounding observations. GPT-5.6 Sol
classifies action purposes, and we manually review the category assignments.
The principal categories are achieving a goal or subgoal, collecting
information to update the world model, and serving both purposes.
Information-seeking purposes further divide into exploring an unfamiliar
situation without an existing hypothesis, testing a proposed mechanism, and
investigating an unexpected outcome. One author independently reviewed 30
action segments; all 30 classifications were correct (100\% accuracy).

\subsubsection{Errors and recovery}
\label{sec:error-recovery-annotation}
\label{sec:explanation-ratings}

We annotate errors throughout both human and LLM trajectories, distinguishing
incorrect mechanism beliefs, planning errors, and execution errors. A mechanism
error is an accepted belief that conflicts with the game's actual behavior;
a tentative hypothesis alone does not establish such an error. Planning failures
are attempts that do not achieve an explicit goal or subgoal, while execution
errors occur when the actual input deviates from the intended operation. We use the game source,
observed transitions, and the learner's statements and actions to establish
the mistake and locate the supporting interval. We also record explicit
expressions of not understanding a rule without a concrete explanation, and
uncertainty about the level goal, which can coexist with candidate goals.
Continuous expressions concerning the same issue are merged, and these
unknowns are distinguished from accepted but incorrect conclusions.

We then link each error to any later evidence of correction. Correcting a
mechanism error requires replacing the same incorrect belief; correcting a
planning or execution error requires addressing the mistaken plan or action.
We retain the evidence for both the error and its correction, including cases
where correction is not observed. Section~\ref{sec:cognitive-errors} reports
event counts and the numbers of observed corrections or subsequent
understanding.

\reviewcomment{A13}


\section{Data analysis}
\label{sec:human-world-models}

Our corpus preserves state changes, actions, and concurrent thought reports
during interactive problem solving, providing rich primary material for
studying how humans understand unfamiliar environments. The following analyses
examine selected cognitive phenomena and compare human trajectories with
those of LLMs to identify similarities and differences. %
These analyses only begin to explore the corpus's research potential.
Researchers can return to the original records to develop and test further
hypotheses about cognition in interactive learning.

\subsection{Humans draw on prior knowledge}
\label{sec:anticipatory-recognition}

We find that humans can infer the existence of a mechanism before it is first
triggered.
Comparing each mechanism's first trigger with its first correct recognition,
we find that 57 of 1,425 paired mechanism instances (4.0\%) are recognized
before their first trigger. %
We interpret this as evidence that prior knowledge helps people anticipate
how an unfamiliar environment works. For example, a participant in TU93
identifies the green endpoint as the goal and moves toward it before reaching
it: recognition is recorded before action 14, while the first trigger occurs
after action 23 (Figure~\ref{fig:prior-tu93-anticipation}).

LLMs also exhibit this property. The proportions of mechanisms recognized
before their first trigger are 11.3\% and 9.6\% for \vista{}--Claude and
\vista{}--GPT, 10.8\% and 9.2\% for Schema--Claude and Schema--GPT, and
7.3\% for OPINE--Claude (Table~\ref{tab:induction-coverage}). Thus, anticipatory
recognition is present in both groups and is more frequent in these LLM
configurations.

People also use analogies to familiar mechanisms to form expectations about
an unfamiliar world model. Analogies to knowledge outside the current game
occur in 37/144 human trajectories (25.7\%). For example,
after an attempted interaction, a BP35 participant reports having interpreted
an object as a Mario-style teleport pipe, then rejects that interpretation
based on the feedback (Figure~\ref{fig:prior-bp35-analogy}). Such analogies
supply an initial account that can be refined through interaction.

LLMs likewise use analogies, in 26/119 successful trajectories (21.8\%):
a \vista{} agent relates LF52 to peg solitaire, while a Schema agent
interprets FT09 through Lights Out. Familiar structures therefore guide
world-model construction in both humans and LLMs.

\subsection{Humans use qualitative representations}
\label{sec:qualitative-world-models}

Humans often form a coarse understanding of a mechanism that leaves some
of its details unspecified. We refer to this as a \emph{qualitative
representation}: the learner expresses how a mechanism works without
specifying all relevant conditions, timing, or magnitudes. For example,
a TU93 participant describes an enemy as following the player's path and
plans to lead it on a detour, without specifying its movement delay.
In G50T, a participant initially describes a gray clone as following them,
without specifying how its replay is coupled to successful player moves.
These examples show how a coarse account of a mechanism can guide further
reasoning and action.

Qualitative descriptions occur in 72 of the 144 human trajectories (50.0\%).
LLMs also use qualitative representations, in 17 of the 119 successful
trajectories (14.3\%). Qualitative
descriptions are therefore more common in human reports.
LLM records also contain precise numerical descriptions. For example,
Schema--GPT explicitly records a six-pixel downward movement in G50T;
OPINE--Claude represents a patrolling enemy's route in TU93 by five row
coordinates, $[16,22,28,34,40]$, and identifies a round-trip period of eight
valid moves.

\subsection{Action purposes}
\label{sec:action-purposes}

We analyze the purposes of human actions and find three principal
categories: achieving a goal or subgoal, collecting information to update
the world model, and serving both purposes. This distinction is consistent
with WorldTest's design, which separates gathering information to learn a
world model from using that model for downstream tasks~\citep{warrier2025worldtest}.
Specifically, among human actions, 47.2\% pursue goals, 9.1\% collect information, and
28.4\% serve both purposes.
The corresponding proportions for successful LLM trajectories are 57.1\%,
11.9\%, and 30.5\%. %

For example, a TU93 participant proposes approaching and eating enemies,
then follows the path and eliminates them, illustrating goal pursuit
(Figure~\ref{fig:purpose-tu93-goal}). In CN04, another participant attempts
to connect shapes while saying they will ``try and see what happens,''
working toward the connection goal while learning which configurations work
(Figure~\ref{fig:purpose-cn04-dual}).

Further analysis shows that exploration serves several purposes. Sometimes
people explore an unfamiliar situation without an existing hypothesis to
find out what happens. For example, an LF52 participant says they do not know
what the newly introduced purple objects are, then clicks them and tries
jumping a green object across them (Figure~\ref{fig:purpose-lf52-unfamiliar}).
Sometimes people act to test a proposed mechanism: an SC25 participant
hypothesizes that a yellow pattern triggers teleportation, draws the pattern,
and checks the resulting position (Figure~\ref{fig:purpose-sc25-hypothesis}).
At other times, people investigate unexpected outcomes. Later in SC25, the
same participant arrives somewhere other than expected, revisits earlier
actions and tries other patterns, then proposes that the character's size
may affect the result (Figure~\ref{fig:purpose-sc25-surprise}).

Among human actions with an information purpose, including those that also
pursue a goal, unfamiliar-situation exploration, hypothesis testing, and
investigation of unexpected outcomes account for 40.6\%, 49.2\%, and 9.8\%,
respectively; 0.1\% combine these situations and 0.3\% lack sufficient evidence
for a subtype. For LLMs, the corresponding proportions are 10.7\%, 49.0\%,
and 4.8\%, with 35.5\% combining situations. Hypothesis testing is common in
both groups, while humans more often explore without a prior hypothesis and
LLM information-seeking segments more often combine these situations.

\subsection{Errors and recovery}
\label{sec:cognitive-errors}

We find that humans make errors while forming world models, planning, and
executing actions, and can revise their understanding and actions through
further interaction. Across 144 human trajectories, we record 15 accepted
conclusions that conflict with the actual mechanisms, with subsequent
corrections observed in 11 cases. For example, in game CN04, a participant
initially believes that a shape must enter another shape before its length
can change. They later discover that clicking its white endpoint is sufficient
and explicitly retract the earlier explanation
(Figure~\ref{fig:recovery-cn04-belief}). We also record 79 occasions on which
participants explicitly express that they do not understand a rule and offer
no concrete explanation, and 57 occasions on which the level goal remains
undetermined. In LF52, a participant repeatedly clicks without understanding
how to move objects along the tracks, then discovers the arrow-key controls
and recognizes that multiple platforms move together
(Figure~\ref{fig:recovery-lf52-control}). Another excerpt from CN04 shows a
participant progressing from uncertainty about how the red endpoints should
meet to understanding how the shapes should fit together
(Figure~\ref{fig:recovery-cn04-goal}).

Human actions also exhibit planning failures and execution deviations. We
record 1002 attempts that fail to achieve an intended goal or subgoal, with
subsequent recovery observed in 916 cases. In LP85, a participant plans to
position the orange object before moving the yellow one, but discovers that
the latter operation also moves the orange object and then reconsiders their
coupled movement (Figure~\ref{fig:recovery-lp85-plan}). We further confirm
136 execution deviations in which the actual input differs from the intended
operation, with 134 subsequently corrected. For example, a participant clicks
outside a control's boundary and, after observing no effect, adjusts the click
position and continues (Figure~\ref{fig:recovery-lp85-input}). These cases
show how unsuccessful plans or input deviations lead to action adjustments
and further progress.

The 119 successful LLM+harness trajectories also contain errors and recovery:
451 of 593 accepted but incorrect mechanism conclusions are subsequently
corrected. We additionally record 29 occasions of explicitly not understanding
a rule without a concrete explanation, and 83 occasions of an undetermined
level goal. For LLMs, goal uncertainty often coexists with concrete candidates:
79 of the 83 occasions include one or more candidate goals, compared with
36 of 57 for humans. For example, Schema--Claude explicitly states that the
goal in AR25 remains unknown and explores alternatives such as aligning a
shape with a target or fitting it into a template, before using interaction
feedback to identify the current level's winning condition. The larger number
of recorded incorrect mechanism conclusions for LLMs may partly reflect how
their reasoning is expressed: public reasoning, code, and notes articulate
mechanism understanding as explicit, checkable claims, making these conclusions
and their revisions easier to observe. Human concurrent verbal reports may
not express every change in understanding.

LLM trajectories also contain 822 attempts that fail to achieve an intended
goal or subgoal, with subsequent recovery in 728 cases, and seven confirmed
execution deviations, six of which are subsequently corrected.

\section{Implications for Benchmark Design}
\label{sec:implications}

Current LLMs, supported by effective harnesses, can perform well on
\arc{} tasks.\reviewcomment{A19} By connecting action trajectories with
concurrent thought reports, our dataset helps characterize how humans build,
use, and revise world models in unfamiliar environments. These observations
inform the design of next-generation benchmarks for interactive world-model
learning.

\label{sec:benchmark-design}

\paragraph{Exploring counterfactual and emergent mechanisms.}
Both humans and LLMs infer some mechanisms before triggering them and use
analogies to familiar concepts to understand the environment
(Section~\ref{sec:anticipatory-recognition}). Such priors allow some mechanisms
to be predicted without elaborate exploration.\reviewcomment{A20} Future
benchmarks could reduce reliance on familiar priors through counterfactual
mechanisms. Mars~\citep{tang2024mars}, for example, changes terrain, character
behavior, and task dependencies to depart from commonsense expectations,
requiring learners to discover the actual rules through interaction.
Another direction is emergent mechanisms: developers specify local dynamics
rather than directly encoding complete higher-level mechanisms, whose effects
arise from interactions among components. Learners would need to construct
and test combinations to discover configurations that produce useful effects.

\paragraph{Induction through accumulated evidence and abstraction.}
Many current \arc{} mechanisms can be understood from local, directly visible
state changes, leaving room for more demanding inference. Among mechanism
instances with both trigger and recognition annotations, the mean action
difference between first recognition and first trigger is 90.8 for humans
and 24.4 for LLMs with harnesses. %
Future benchmarks could draw on AutumnBench's hidden states and stochastic
outcomes~\citep{warrier2025worldtest}, and introduce delayed effects to require
evidence accumulated across interventions and inference about unobserved
causal relationships. Humans also frequently guide actions with qualitative
representations that omit precise details, more often than LLMs do
(Section~\ref{sec:qualitative-world-models}). The big world hypothesis emphasizes
that an environment's complexity can far exceed an agent's finite memory and
computation~\citep{javed2024bigworld}. Benchmarks could accordingly make complete,
precise pixel-level modeling infeasible, requiring qualitative abstractions
that serve current goals and posing new challenges for LLM world-model learning.

\paragraph{Planning useful subgoals and testing combined outcomes.}
Future \arc{} tasks could require longer plans and deeper subgoal dependencies.
After recognizing all relevant main mechanisms, humans take 95.1 further actions
on average to complete eligible levels, versus 49.2 for LLMs with harnesses.
Constructing a device such as an automated sorting and transport system would
pose two challenges. First, learners must derive useful subgoals from the
intended function, identifying and building the required components. Second,
even known local mechanisms may have uncertain combined effects, requiring
learners to test the assembled device and revise their plans. Planning would
thus combine subgoal construction with experimentation.\reviewcomment{A21}

\section{Conclusion}
\label{sec:conclusion}

We introduce a dataset of human think-aloud reports aligned with gameplay on
\arc{}, paired with LLM trajectories and annotations of world-model changes,
action purposes, and errors and recovery. Our analyses highlight how learners
use prior knowledge, reason with qualitative representations, explore, and
revise their understanding and actions. These records support further research
on interactive learning and motivate next-generation benchmarks that challenge
familiar priors and require cumulative learning, abstraction, and planning.

\subsection*{Ethics Statement}

This study was approved by an institutional review board (details withheld
for anonymous review). All participants provided informed consent and were
compensated for their time through Prolific.

\bibliography{references}
\bibliographystyle{iclr2027_conference}

\clearpage
\appendix
\section{Human Data Collection Protocol}
\label{app:human-protocol}

The live-interaction interface introduces the study before presenting a
controls tutorial and a separate practice game. Participants provide informed
consent and enable their microphone; the microphone check asks them to say a
sentence and detects speech before continuing. The study overview presents
the following instructions:
\begin{participantinstructions}{Study instructions}
You will play 3 different video games. The goal of each game is unknown.
You need to experiment, infer what the goal might be, and speak your thought
process into the microphone at the same time.

Try to win each game while thinking aloud continuously. Say everything that
is going through your mind as you play.

We care primarily about how detailed your think-aloud explanation is---not
how quickly you finish. We estimate about 20 minutes per game, but it is fine
to take longer; you will be compensated for the additional time.
\end{participantinstructions}

The controls tutorial explains that available controls vary by game and
unavailable controls are disabled. It introduces directional input using
on-screen buttons or arrow keys, the space action, clicking directly on
objects or positions, undo when supported, and resetting the current level.
It notes that the practice game does not support the space or undo actions.
These instructions explain interface operations without supplying the mechanisms
or solutions of the formal games. Before practice, participants are told:
\begin{participantinstructions}{Practice instructions}
You will now have 10 minutes to beat one practice game. Keep thinking aloud
from start to finish---say everything that is going through your mind.
Your microphone will record continuously.
\end{participantinstructions}

Practice has a 10-minute limit, and completing the practice game unlocks the
formal study. Giving up or reaching the practice deadline without completing
the game ends participation. The formal study assigns three games by default,
each with a 60-minute maximum, and includes the games' full level sequences.
The 20-minute duration in the overview is an estimate communicated to
participants, not the enforced time limit. Giving up a formal game or reaching
its time limit ends the session without offering the remaining games.

During both practice and formal play, approximately 10 seconds without
detected speech triggers a reminder to continue verbalizing current thoughts.
The reminder disappears when speech resumes and provides no game-specific
advice. The timer excludes time spent in a background browser tab, and
reminders are suppressed during instructional overlays and terminal states.
The instructions communicate compensation for additional time; the exit
message for Prolific participants explains that compensation for submitted
participation is paid through a Prolific bonus after returning the study.

The reported 144 trajectories constitute the human analysis corpus across
the 25 public games, rather than all assigned attempts or practice runs.
As described in Section~\ref{sec:data-collection}, the human analysis
uses completed games. The default assignment of three games therefore does
not imply three retained trajectories per participant.

\section{Annotation procedures and review details}
\label{app:annotation-details}

\subsection{Mechanism catalog}
\label{app:annotation-catalog}

Using the definition in Section~\ref{sec:preliminaries}, we catalog mechanisms
that can independently change a player's prediction, action choice, or risk assessment. For example, clicking a green
support removes it in Table~\ref{tab:mechanism-examples}. For every game and
level, two authors inspect the game source, rendered states, and representative
replays and jointly catalog the key mechanisms through discussion. We split an
engine event when it produces multiple independently meaningful effects, and
exclude rendering details, cosmetic changes, one-off solution steps, and the
mere reuse of an earlier mechanism in a harder layout.  Each mechanism receives a
stable identifier, a natural-language definition, its earliest observable
level, and a scope label.  The primary analysis uses \emph{main} mechanisms;
optional failure, undo, or resource-exhaustion behaviors are retained as
supplemental mechanisms. We denote the mechanism set for game $g$ by
$\mathcal M_g$.

\subsection{Trigger detection}
\label{app:annotation-triggers}

We insert mechanism-specific trigger-detection code into the ARC game source
and execute the instrumented game during trajectory replay. The detection code
checks changes in selected internal variables before and after each action,
as well as whether relevant code paths execute. We use an LLM as a
code-generation assistant to turn each mechanism definition into a candidate
runtime monitor. The LLM is given the relevant game source,
the mechanism description, and representative before/after states.  It
proposes checks over state changes, actions, and distinctive source paths.
Annotators review the proposed code, run it against positive and negative
replays, and revise it until the event has a precise player-level meaning.  A
validated monitor has the form
\begin{equation}
  d_m(s_{a-1},u_a,s_a,c_a)\in\{0,1\},\qquad
  t^x_{\tau m}=\min\{a:d_m=1\},
  \label{eq:monitor}
\end{equation}
where $u_a$ is the $a$th player action, $s_{a-1}$ and $s_a$ are the game states
before and after that action, including the monitored internal variables,
$c_a$ records whether the relevant code paths executed, and $t^x_{\tau m}$ is the first
action at which mechanism $m$ is objectively triggered in trajectory $\tau$.
The LLM proposes monitor code;
the accepted mechanism definition, binding, and test result are human
verified.  Monitor activation records objective execution of a known mechanism and
does not label whether the player understood it.
We manually inspected 30 trigger annotations and found all 30 to be correct.

\subsection{Recognition evidence and review coverage}
\label{app:annotation-recognition}

Recognition timing is annotated from the full chronological trajectory
context. Agent and human records are annotated separately with
LLM assistance to preserve their different evidence modalities. Agent inputs
contain the mechanism definition, public outputs and any released reasoning,
harness messages, aligned actions, and visible outcomes. Human inputs contain
the same definition, aligned think-aloud transcript, actions, and outcomes.
For each trajectory--mechanism pair, $t^r_{\tau m}$ denotes the first action
with sufficient evidence of recognizing mechanism $m$. The annotation also
records supporting text or behavior, earlier partial or incorrect beliefs,
and later corrections or drift.

We accept two kinds of recognition evidence.  \emph{Explicit evidence} is a
public statement that correctly specifies the relevant causal mechanism,
condition, direction, object, or goal.  \emph{Strong behavioral evidence} is a
purposeful prediction, diagnostic action, or plan followed by an outcome that
rules out live alternative explanations.  Accidental triggering, ordinary
movement, a forced route, or level completion alone is insufficient.  The
annotation distinguishes recognized, partial-only, available-but-not-
recognized, not-yet-observable, and unreviewable cases.

Recognition review checks the quoted evidence against the original log and
action context, using the frozen game source and relevant replay states to
assess whether the stated mechanism is correct. When a human verbal reference or
visual consequence is ambiguous, we consult the aligned replay to interpret
the report in its action context. LLM annotators review
human and agent records chronologically, checking the available verbal or
reasoning evidence against the action histories, game source, and replay
states. Labels are not assigned from keyword matches or successful completion
alone. We manually checked 100 annotations, of which 97 were correct.
Each decision identifies exact evidence, its action boundary, and later
retention or drift. Source code establishes the environmental mechanism, not
whether the acting model knew it. The default analysis uses the adjudicated
best-evidence timestamp; explicit-only and behavior-only timestamps are
retained for sensitivity checks where separately supported.

\subsection{Action purposes, errors, and recovery}
\label{app:annotation-behavior}

Following Section~\ref{sec:action-purpose-annotation}, we classify contiguous
human and LLM action segments as goal pursuit, information gathering, or both.
Information gathering includes exploring unfamiliar situations, testing
hypotheses, and investigating unexpected outcomes. Labels use the aligned
verbal or reasoning evidence and surrounding observations. GPT-5.6 Sol assists
classification; one author reviewed 30 segments and found all 30 correct.

Error annotations distinguish accepted but incorrect mechanism beliefs,
attempts that fail to achieve an explicit goal or subgoal, and inputs that
deviate from the intended operation (Section~\ref{sec:error-recovery-annotation}).
Tentative hypotheses, explicitly not understanding a rule, and uncertainty
about the level goal are distinguished from accepted incorrect conclusions.
Continuous expressions about the same issue are merged into one event.

Each error is linked to later evidence that corrects the same belief, plan,
or input; cases without observed correction are retained. Annotations preserve
supporting statements, actions, and outcomes, with the game source and replay
establishing the relevant mechanism. Section~\ref{sec:cognitive-errors} reports
event counts and observed corrections.

\clearpage
\section{Human trajectories and concurrent verbal reports}
\label{app:human-trajectory-examples}

These excerpts show humans drawing on prior knowledge, pursuing goals,
gathering information, revising their understanding, and adjusting their actions.
Game frames and original verbal
reports are aligned independently to the action axis; pale bands and connectors
locate the estimated speech intervals. A checkmark indicates level completion
in place of the automatic next-level observation. Horizontal spacing is
schematic, and ellipses indicate omitted actions.

\subsection{Prior knowledge}
\label{app:human-prior-examples}

These excerpts illustrate the two human cases in
Section~\ref{sec:anticipatory-recognition}.

\begin{figure}[!htbp]
  \centering
  \includegraphics[width=\linewidth]{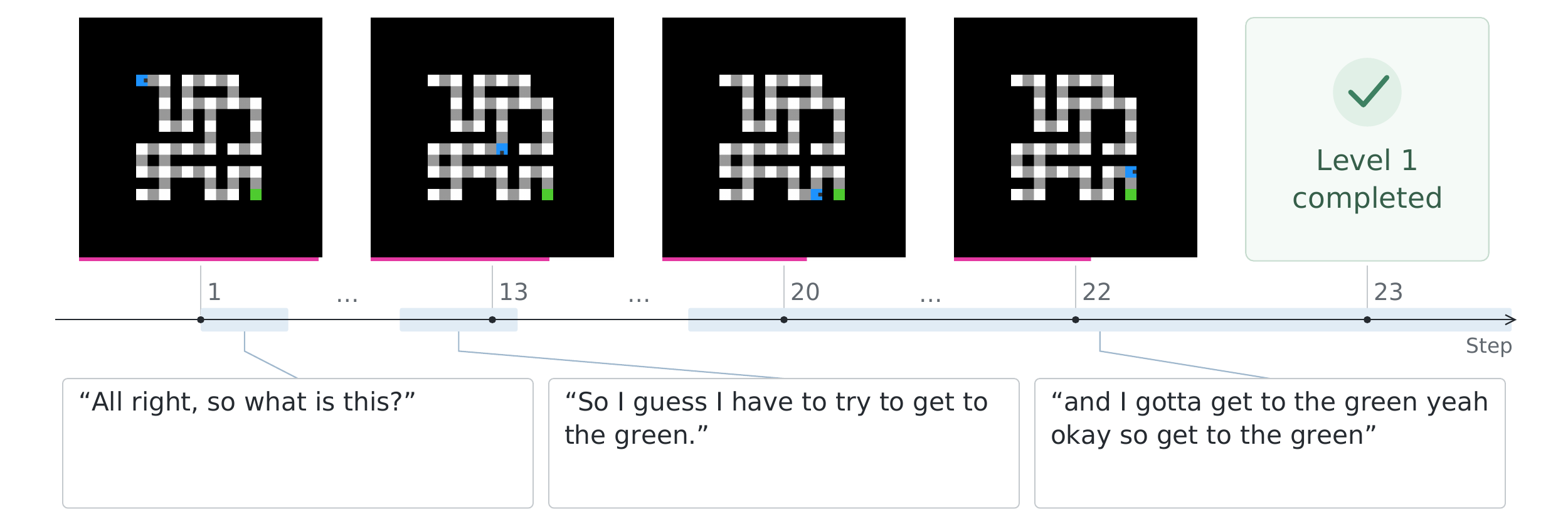}
  \caption{\textbf{Inferring a goal before reaching it (TU93).} The participant identifies the green location as a candidate goal before first reaching it.}
  \label{fig:prior-tu93-anticipation}
\end{figure}

\begin{figure}[!htbp]
  \centering
  \includegraphics[width=\linewidth]{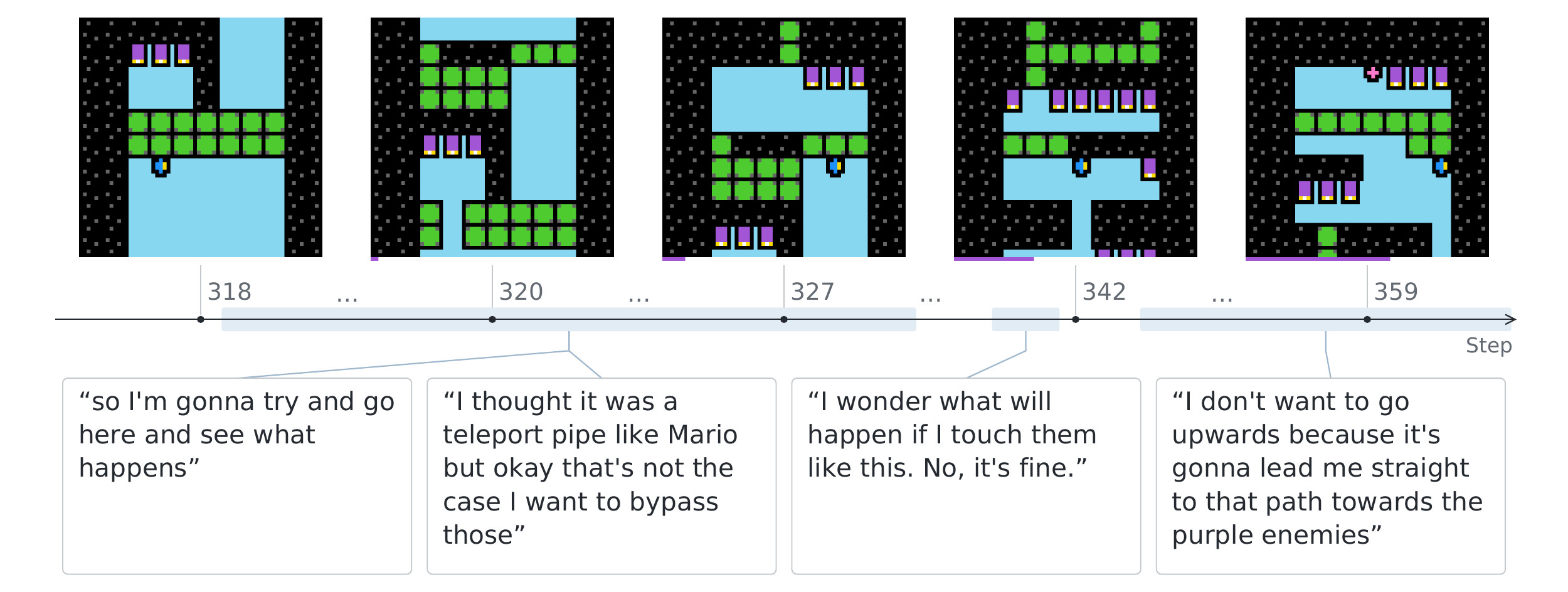}
  \caption{\textbf{Revising an analogy-based hypothesis (BP35).} The participant retrospectively describes a Mario-style teleport-pipe interpretation, rejects it, and tries another route.}
  \label{fig:prior-bp35-analogy}
\end{figure}

\clearpage

\subsection{Action purposes}
\label{app:human-action-purpose-examples}

These excerpts illustrate goal pursuit, information gathering, and their
combination in Section~\ref{sec:action-purposes}. The information-gathering
examples show exploration of unfamiliar objects, hypothesis testing, and
investigation prompted by an unexpected outcome.

\begin{figure}[!htbp]
  \centering
  \includegraphics[width=\linewidth]{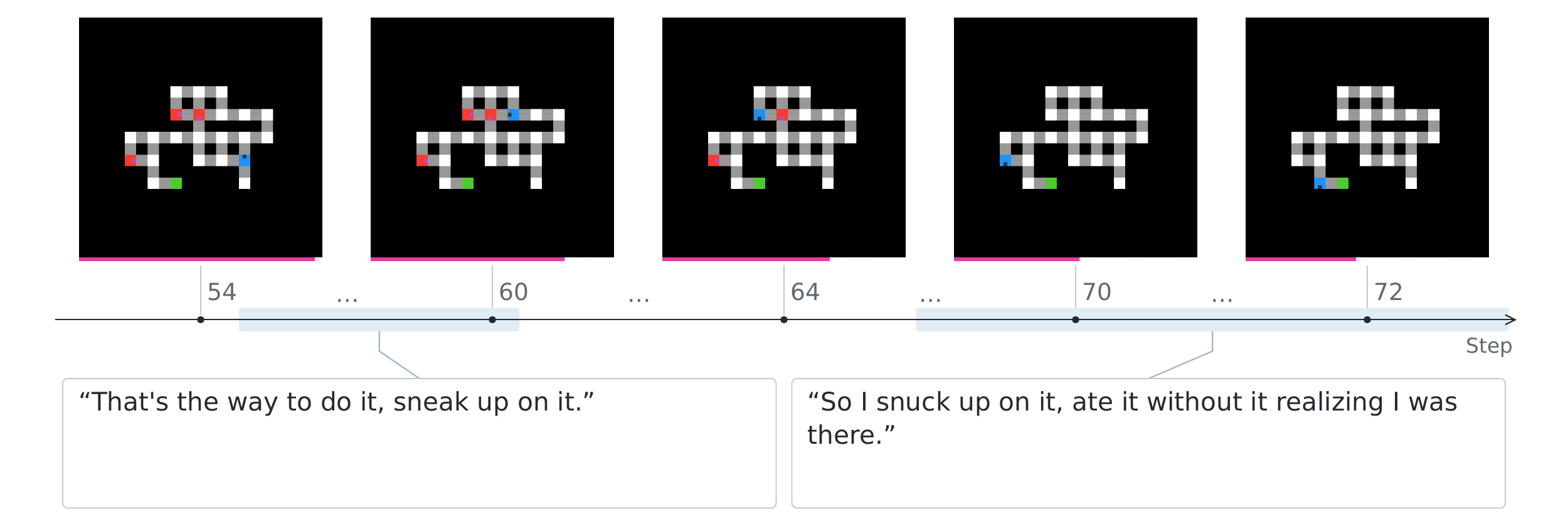}
  \caption{\textbf{Pursuing a subgoal (TU93).} The participant plans an approach and executes a sequence that removes the enemies.}
  \label{fig:purpose-tu93-goal}
\end{figure}

\begin{figure}[!htbp]
  \centering
  \includegraphics[width=\linewidth]{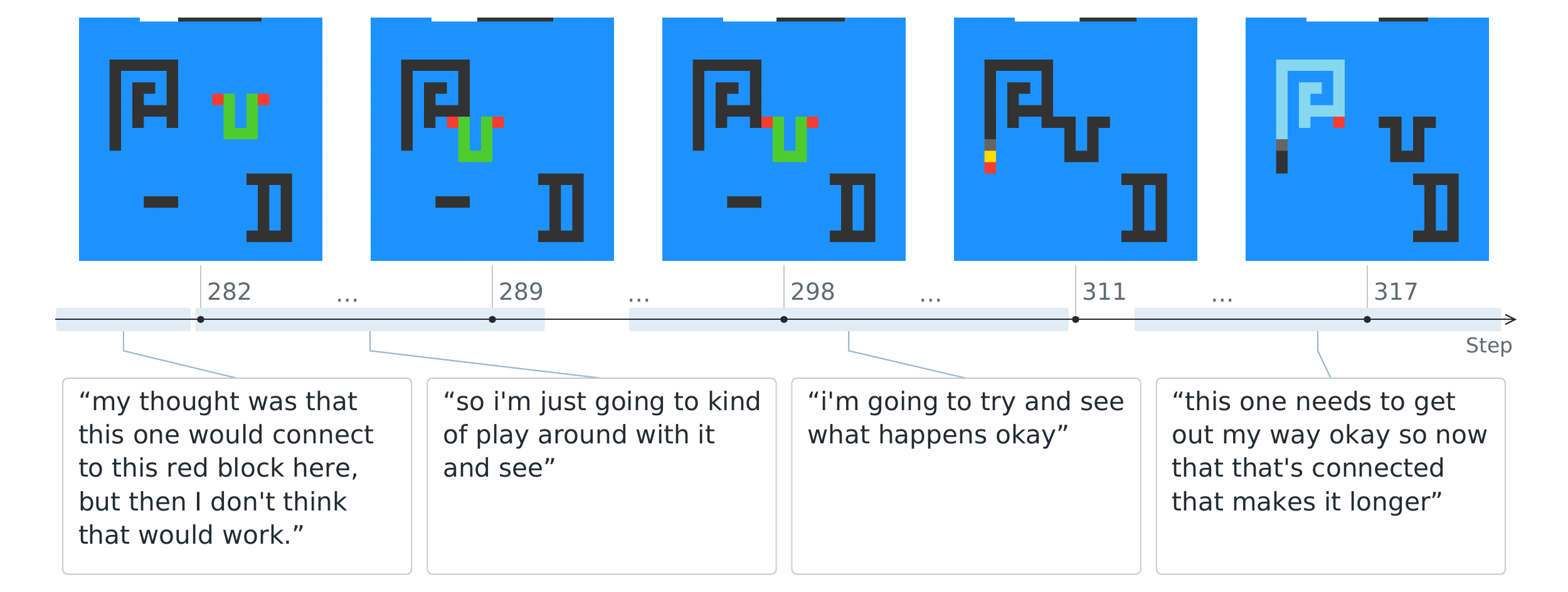}
  \caption{\textbf{Combining goal pursuit and information gathering (CN04).} Testing candidate shape connections both advances the solution and reveals which configurations work.}
  \label{fig:purpose-cn04-dual}
\end{figure}

\begin{figure}[!htbp]
  \centering
  \includegraphics[width=\linewidth]{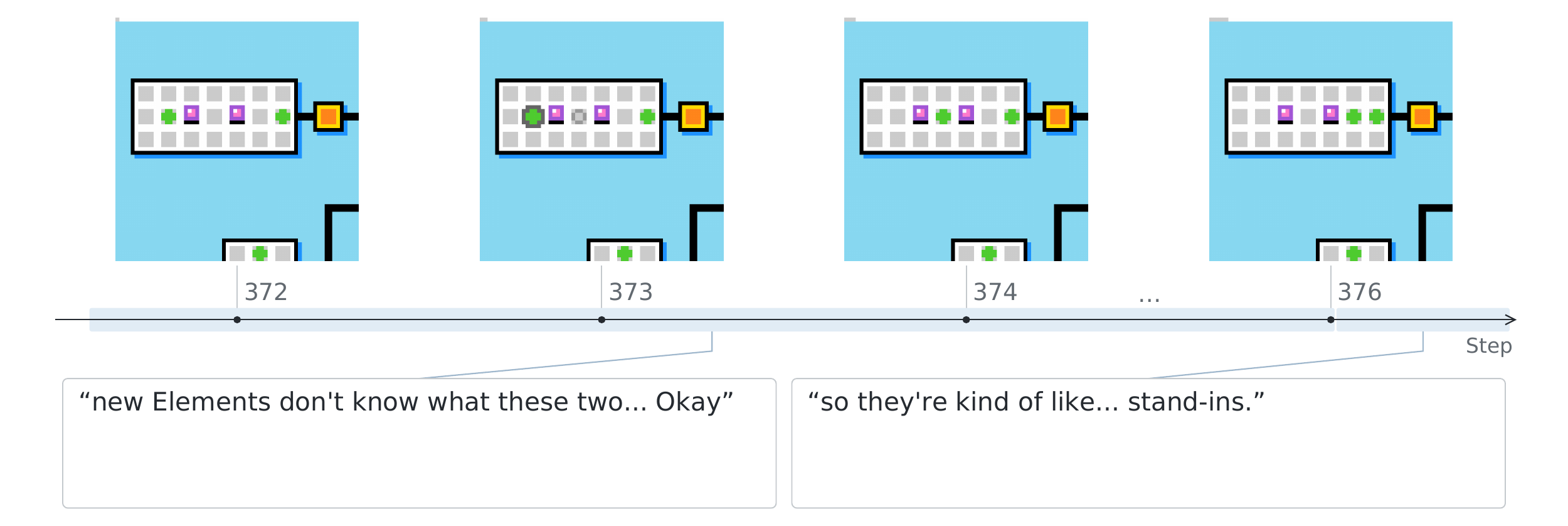}
  \caption{\textbf{Exploring unfamiliar objects (LF52).} The participant tests clicks and movement before proposing an explanation for the purple objects.}
  \label{fig:purpose-lf52-unfamiliar}
\end{figure}

\begin{figure}[!htbp]
  \centering
  \includegraphics[width=\linewidth]{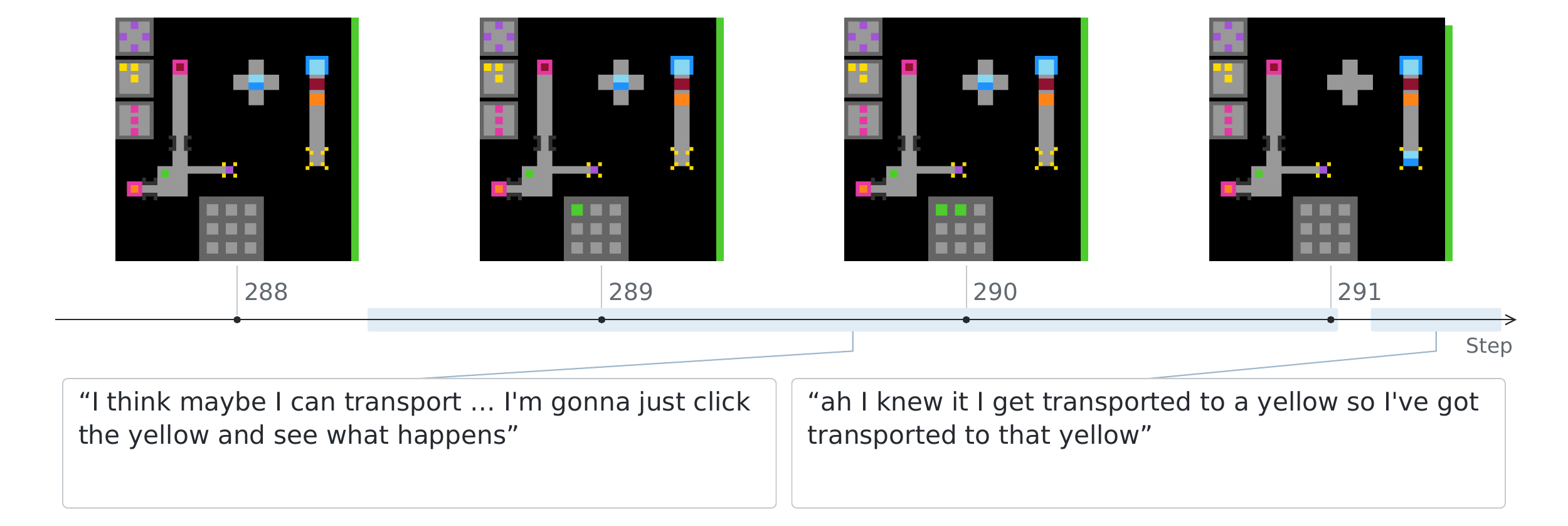}
  \caption{\textbf{Testing a hypothesis (SC25).} The participant predicts teleportation, tests the yellow pattern, and confirms the predicted effect.}
  \label{fig:purpose-sc25-hypothesis}
\end{figure}

\begin{figure}[!htbp]
  \centering
  \includegraphics[width=\linewidth]{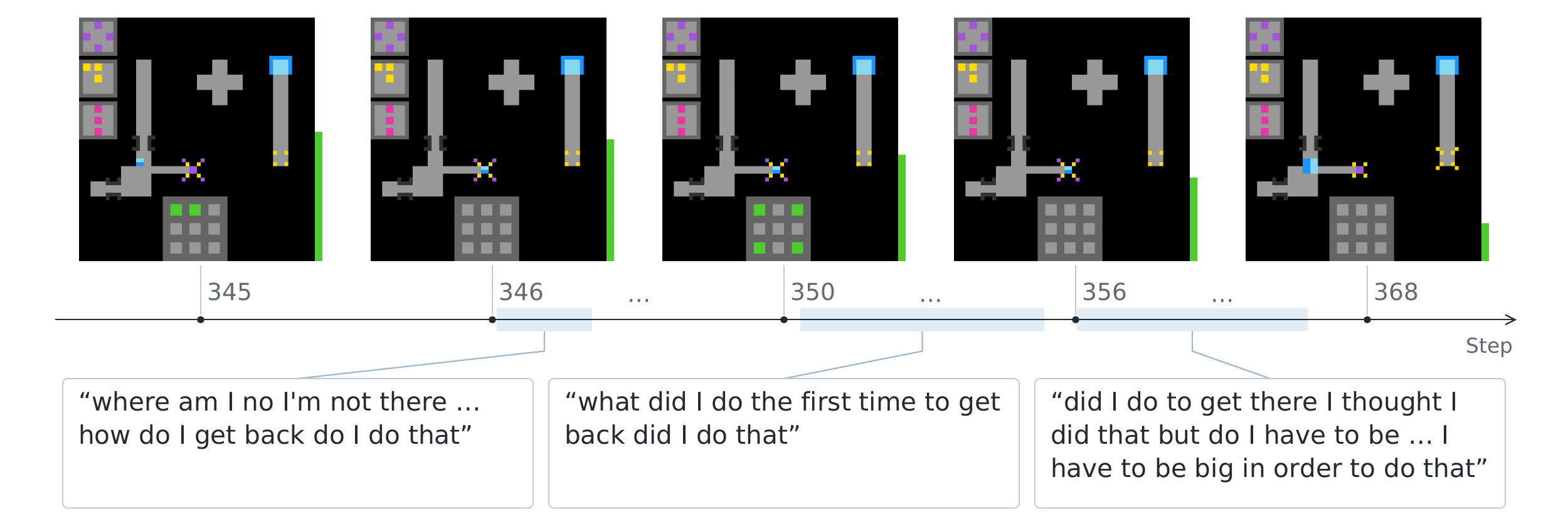}
  \caption{\textbf{Investigating a prediction error (SC25).} An unexpected teleport destination prompts further tests and the hypothesis that returning requires a larger character.}
  \label{fig:purpose-sc25-surprise}
\end{figure}

\clearpage

\subsection{Revised understanding and action}
\label{app:human-recovery-examples}

The following excerpts illustrate the human cases discussed in
Section~\ref{sec:cognitive-errors}.

\begin{figure}[!htbp]
  \centering
  \includegraphics[width=\linewidth]{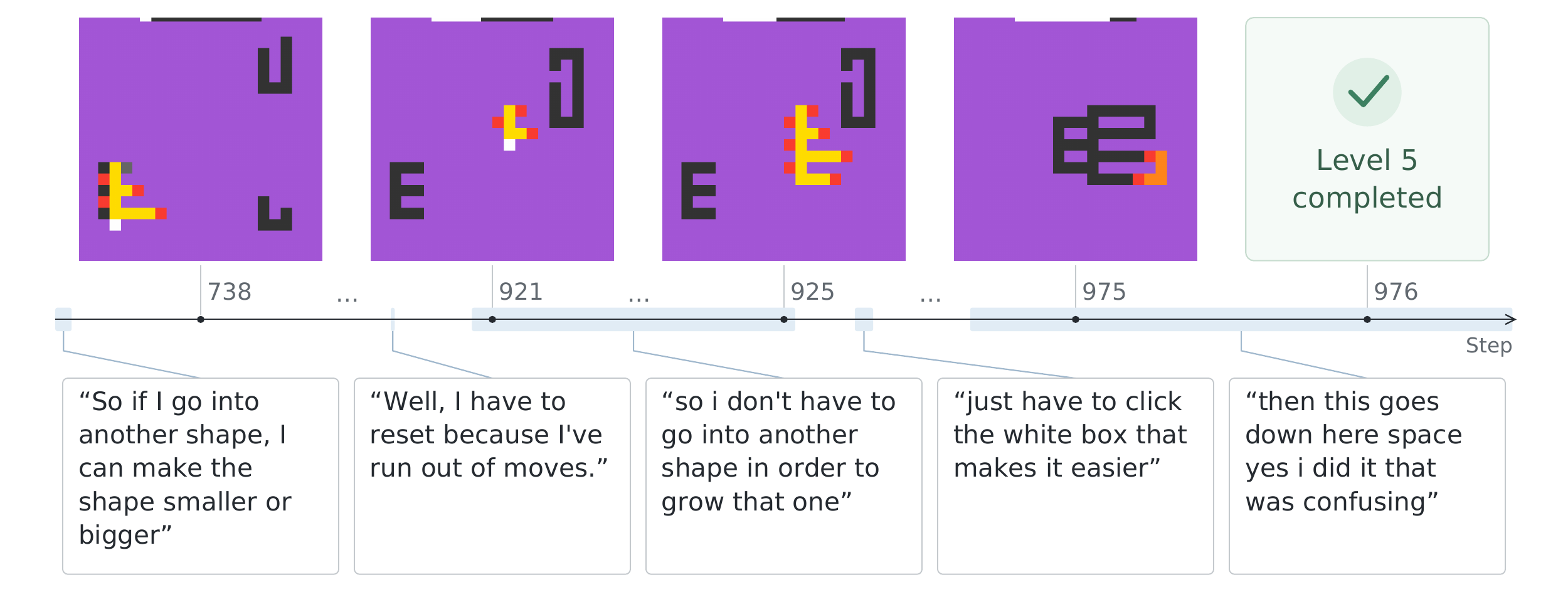}
  \caption{\textbf{Revising a mechanism condition (CN04).} Resizing in place leads the participant to retract the belief that entering another shape is necessary.}
  \label{fig:recovery-cn04-belief}
\end{figure}

\begin{figure}[!htbp]
  \centering
  \includegraphics[width=\linewidth]{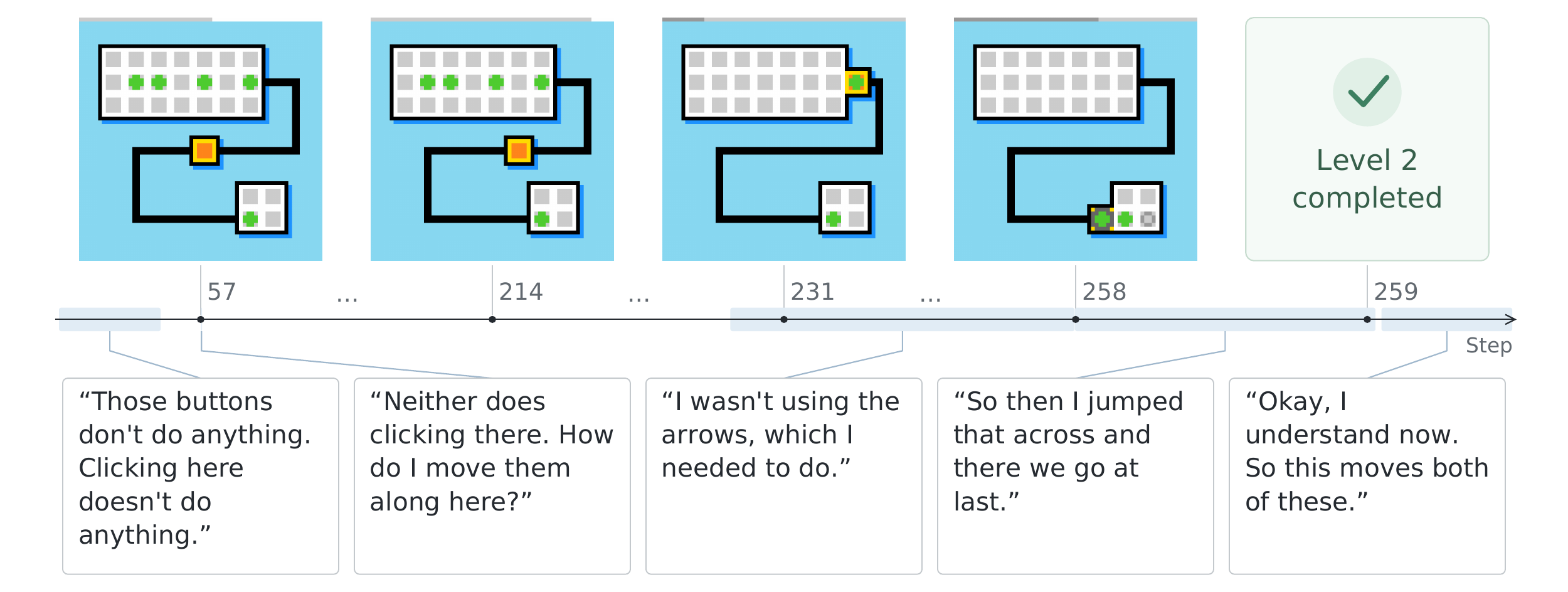}
  \caption{\textbf{Discovering the controls (LF52).} After unsuccessful clicks, the participant identifies the arrow-key controls and completes the level.}
  \label{fig:recovery-lf52-control}
\end{figure}

\begin{figure}[!htbp]
  \centering
  \includegraphics[width=\linewidth]{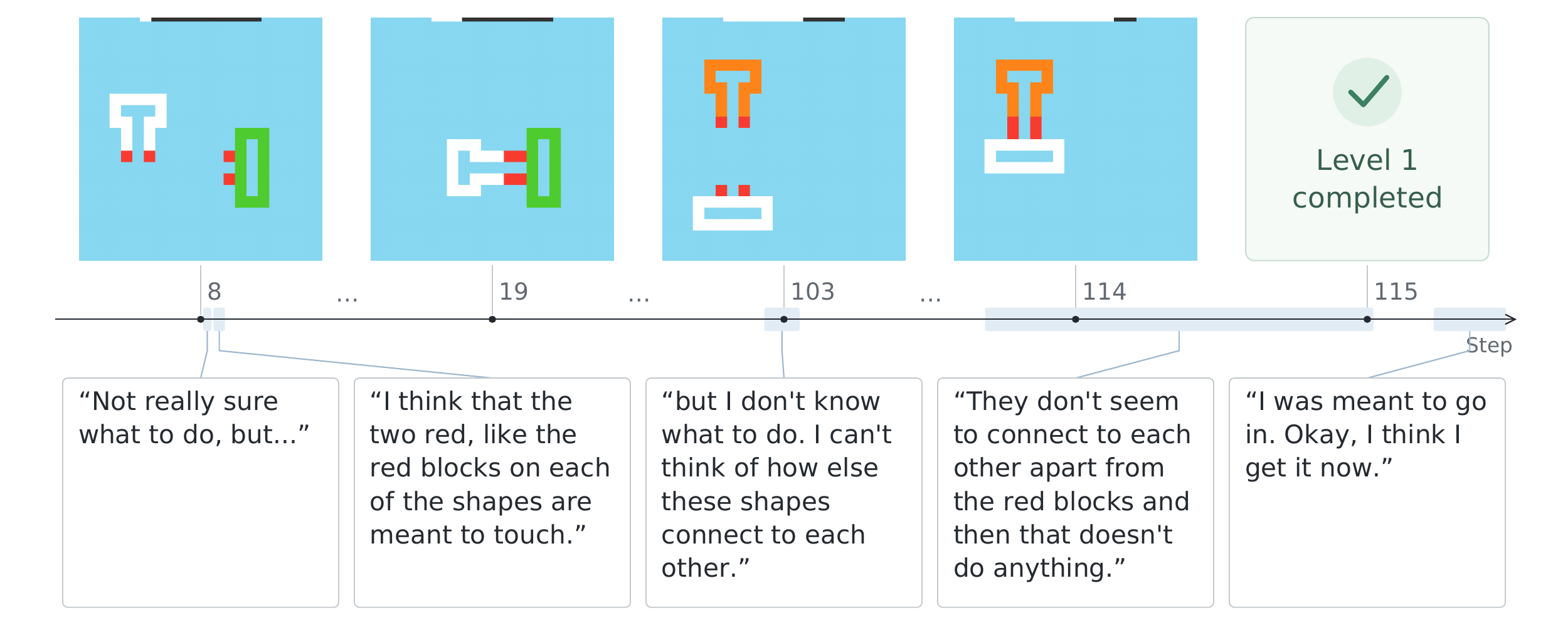}
  \caption{\textbf{Clarifying the connection goal (CN04).} The participant tests how the shapes should fit together and reports understanding after completing the level.}
  \label{fig:recovery-cn04-goal}
\end{figure}

\begin{figure}[!htbp]
  \centering
  \includegraphics[width=\linewidth]{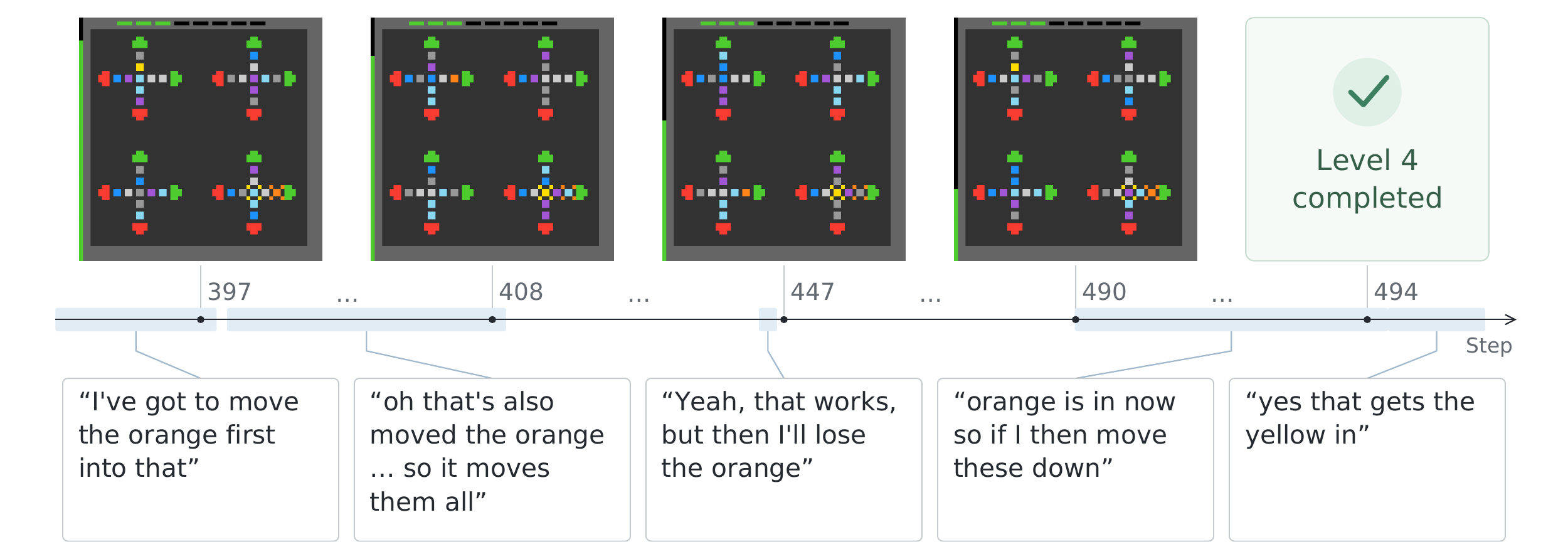}
  \caption{\textbf{Replanning under coupled movement (LP85).} Moving the yellow object disrupts the orange object's placement, prompting a revised sequence that places both at their targets.}
  \label{fig:recovery-lp85-plan}
\end{figure}

\begin{figure}[!htbp]
  \centering
  \includegraphics[width=\linewidth]{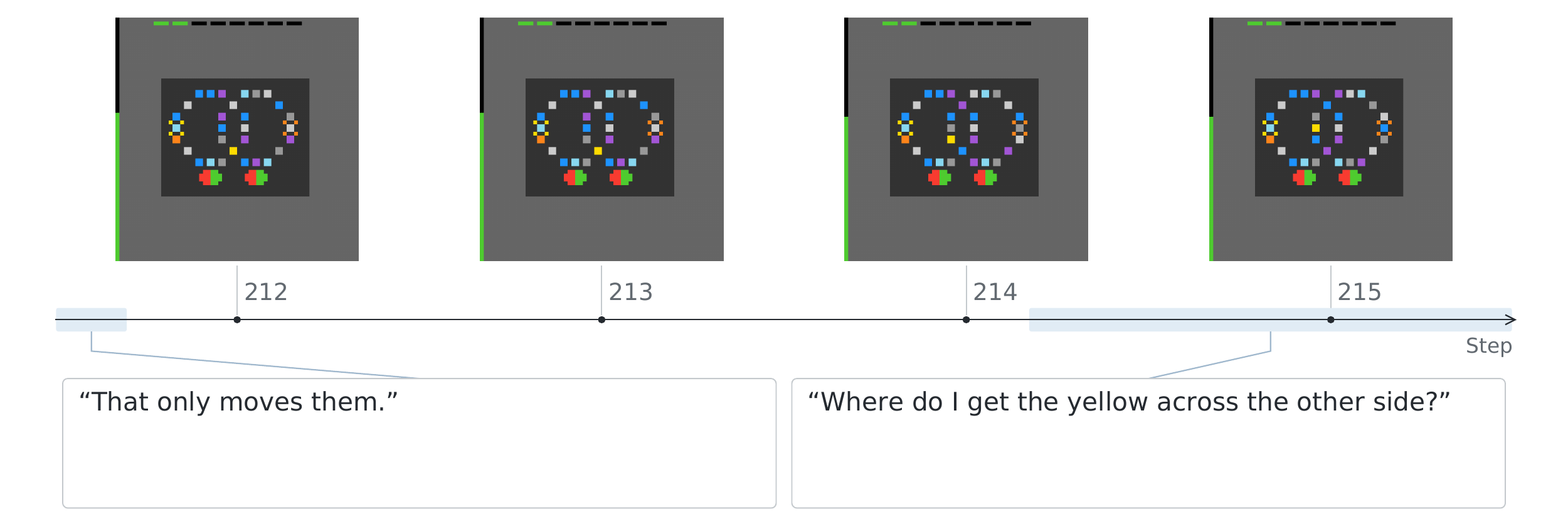}
  \caption{\textbf{Correcting an input (LP85).} A click just outside the button has no effect; a small adjustment activates it.}
  \label{fig:recovery-lp85-input}
\end{figure}

\clearpage

\section{Additional experimental details and results}
\label{app:experiment-details}

\subsection{Systems and comparison scope}

The 394 trajectories contain 236,806 player actions. All conditions use the
same catalog of 354 mechanisms (320 main and 34 supplemental), yielding
5,565 trajectory--mechanism annotation instances.

Direct-agent configurations are Claude Opus 5 high, GPT-5.6 Sol max and high,
and GPT-6 Astra max in the official Standard configuration. \vista{} uses
Claude Opus 5.0 or GPT-5.6 Sol max. Schema--Claude includes 14 Opus 4.8 and
11 Fable 5 runs, all at max effort; Schema--GPT includes 16 GPT-5.6 Sol xhigh
and nine max runs. OPINE-World uses Claude Opus 4.8 high. Astra Standard and
Provider Adapter use the same released model at max effort. Standard uses
the benchmark's direct-agent interface with carried notes, whereas Adapter
preserves additional reasoning state and supports conversation compaction.
These system labels retain the configurations in the public releases.

Human and \vista{} trajectories are selected for successful game completion;
other conditions retain unsuccessful trajectories from the chosen releases.
Model configuration, inference budgets, and the visibility of reasoning
records are not controlled across systems. Comparisons are therefore
descriptive rather than causal estimates of adding a harness. An absence of
public recognition evidence does not establish an absence of internal mechanism
knowledge. The six unsuccessful harness traces consist of one Schema--GPT
and five OPINE--Claude trajectories; they do not estimate a general harness
failure rate.

\subsection{Coverage and conditional action distances}
\label{app:process-coverage}

Tables~\ref{tab:exploration-coverage}--\ref{tab:planning-coverage} provide the
coverage and denominators for the recognition-timing comparisons in
Section~\ref{sec:anticipatory-recognition} and action distances discussed in
Section~\ref{sec:implications}.
Observable instances are main mechanisms introduced at levels the trajectory
has reached. The first action within a level has exploration distance one.
Means and medians use uncapped action coordinates. A mechanism without a
trigger or supported recognition is excluded from the corresponding distance
rather than assigned zero. These measures describe overlapping aspects of
interaction, not mutually exclusive cognitive phases.

\begin{table}[htbp]
\caption{Trigger coverage and supplementary exploration statistics. Observable instances are main mechanisms introduced at reached levels.}
\label{tab:exploration-coverage}
\centering
\small
\begin{tabular}{@{}lrrr@{}}
\toprule
Condition & Triggered/obs. & Coverage & $D^{\rm exp}\leq5$ \\
\midrule
Claude direct & 218/238 & 91.6\% & 42.2\% \\
GPT-max direct & 177/195 & 90.8\% & 46.9\% \\
GPT-high direct & 159/182 & 87.4\% & 39.6\% \\
Astra Standard & 266/285 & 93.3\% & 47.7\% \\
\vista{}--Claude & 294/320 & 91.9\% & 54.1\% \\
\vista{}--GPT & 305/320 & 95.3\% & 49.5\% \\
Schema--Claude & 301/320 & 94.1\% & 45.8\% \\
Schema--GPT & 296/320 & 92.5\% & 51.7\% \\
OPINE--Claude & 285/302 & 94.4\% & 42.5\% \\
Astra Adapter & 287/320 & 89.7\% & 55.7\% \\
Human & 1795/1825 & 98.4\% & 34.1\% \\
\bottomrule
\end{tabular}
\end{table}

Humans trigger 1,795 of 1,825 observable main-mechanism instances (98.4\%);
agent coverage ranges from 87.4\% to 95.3\%. Because the human records are
completion-selected, this is not an exploration success rate for all
recruited participants. Conditional distances can favor a system that reaches
easy mechanisms but misses difficult ones, so timing and coverage answer
different questions.

\begin{table}[htbp]
\caption{Recognition coverage, paired instances, and recognition preceding a trigger.}
\label{tab:induction-coverage}
\centering
\small
\begin{tabular}{@{}lrrr@{}}
\toprule
Condition & Recognition rate & Paired/obs. & $D^{\rm ind}<0$ \\
\midrule
Claude direct & 84.0\% & 193/238 & 11.4\% \\
GPT-max direct & 79.0\% & 150/195 & 12.7\% \\
GPT-high direct & 63.7\% & 114/182 & 13.2\% \\
Astra Standard & 77.5\% & 214/285 & 9.3\% \\
\vista{}--Claude & 92.8\% & 284/320 & 11.3\% \\
\vista{}--GPT & 93.1\% & 291/320 & 9.6\% \\
Schema--Claude & 89.4\% & 278/320 & 10.8\% \\
Schema--GPT & 87.8\% & 271/320 & 9.2\% \\
OPINE--Claude & 83.8\% & 247/302 & 7.3\% \\
Astra Adapter & 41.2\% & 126/320 & 5.6\% \\
Human & 78.1\% & 1425/1825 & 4.0\% \\
\bottomrule
\end{tabular}
\end{table}

Human induction distances use 1,425 instances with both trigger and
recognition annotations. Recognition coverage ranges from 63.7\% to 84.0\%
for direct agents and from 41.2\% to 93.1\% for harness systems, compared
with 78.1\% for humans. Astra Adapter has supported recognition for 41.2\%
of observable mechanisms, versus 77.5\% for Standard, despite completing
25 games versus Standard's 12. This illustrates why observed recognition
coverage should not be equated with task competence: successful behavior may
coexist with sparse mechanism-specific evidence in public records.

Recognition precedes the first monitored trigger in 5.6--13.2\% of paired
agent instances and 4.0\% of human instances. Such cases can reflect inference
from the initial scene, related events, or knowledge carried across levels;
they do not by themselves establish memorization or pretraining exposure.
Trigger definitions are mechanism-specific, so these signed action distances
are not uniform reaction latencies. Evidence before and after an action
shares its integer coordinate, with phase recorded separately.

Later drift is recorded in 55/221 recognized main-mechanism instances for
Astra Standard, 4/132 for Adapter, 10/286 and 11/281 for the two Schema
conditions, and 11/253 for OPINE. Both direct and harness-assisted models can
thus abandon a previously supported mechanism. Differences in public evidence
limit comparisons of drift frequency across systems.

\begin{table}[htbp]
\caption{Eligible planning levels and missing recognition evidence.}
\label{tab:planning-coverage}
\centering
\small
\begin{tabular}{@{}lrr@{}}
\toprule
Condition & Eligible/completed & Missing recognition \\
\midrule
Claude direct & 45/95 & 50 \\
GPT-max direct & 21/48 & 27 \\
GPT-high direct & 10/30 & 20 \\
Astra Standard & 14/137 & 123 \\
\vista{}--Claude & 70/183 & 113 \\
\vista{}--GPT & 79/183 & 104 \\
Schema--Claude & 61/183 & 122 \\
Schema--GPT & 59/182 & 123 \\
OPINE--Claude & 25/160 & 135 \\
Astra Adapter & 1/183 & 182 \\
Human & 228/1057 & 829 \\
\bottomrule
\end{tabular}
\end{table}

Planning distances use completed levels for which every main mechanism
introduced by that level has recognition evidence by completion. Of 1,057
completed human levels, 228 qualify. Astra Standard has 14 qualifying levels
out of 137 completed, and Adapter has one out of 183. Direct GPT-high's mean
of 7.7 actions is based on ten completed levels, despite no full-game wins.
These distances cannot characterize planning on levels without sufficient
recognition evidence. Evidence first recorded after entry into the next level
cannot qualify the preceding level, even when both share an action boundary.
The interval does not isolate all planning activity or reconstruct later
belief drift.


\end{document}